\documentclass[runningheads]{llncs}
\usepackage{graphicx}
\usepackage{amsmath}
\usepackage{orcidlink}
\usepackage{listings}
\usepackage{float} 
\usepackage{caption}
\usepackage{subcaption}
\usepackage{caption}
\usepackage{url}
\usepackage{lscape}
\usepackage[table,xcdraw]{xcolor}
\usepackage{longtable}
\usepackage{cite}
\usepackage{booktabs}
\usepackage{breqn}
\usepackage{comment}
\usepackage{svg}

\usepackage[most]{tcolorbox}

\begin{document}

\makeatletter
\renewenvironment{abstract}{%
      \list{}{\advance\topsep by0.15cm\relax\small
      \leftmargin=0.4cm
      \labelwidth=\z@
      \listparindent=\z@
      \itemindent\listparindent
      \rightmargin\leftmargin}\item[\hskip\labelsep
                                    \bfseries\abstractname]}
    {\endlist}
\makeatother

\title{LLMs as Post-hoc Auditors of Physiological Plausibility in Symbolic Regression: A Clinician-Evaluated Case Study}
\titlerunning{ }

\author{Jorge López-Varela \orcidID{0009-0005-6868-0019}\inst{1, 10} \and 
J. Ignacio Hidalgo\inst{1,10}\orcidID{0000-0002-3046-6368} \and 
José-Manuel Muñoz\inst{2}\orcidID{0009-0007-7134-8708}  \and  
Omar Costilla-Reyes \inst{6}\orcidID{ 0000-0001-8331-7262 } \and
Esther Maqueda MD \inst{3}\and
Jesus Moreno-Fernandez, MD\inst{4,5} \orcidID{0000-0002-9703-1094} \and
Tomás González-Vidal MD\inst{7,8,9}\and
J. Manuel Velasco \inst{1,10}\and
Oscar Garnica \inst{1,10} \orcidID{0000-0001-5064-2587}}

 \authorrunning{López-Varela et al.}

\institute{\fontsize{8}{9}\selectfont
Computer Architecture Department, Universidad Complutense de Madrid, Spain. \email{jorgsa15@ucm.es,hidalgo@ucm.es,mvelascc@ucm.es,ogarnica@ucm.es}\and 
Universidad Autónoma de Baja California, México  \and 
Endocrinology and Nutrition Department. Hospital Universitario de Toledo, Spain \and
Endocrinology and Nutrition Department. Ciudad Real General University Hospital.
Obispo Rafael Torija St. 13005. Ciudad Real. Spain. \and
Castilla-La Mancha Health Research Institute (IDISCAM). Toledo. Spain \and
Equ Healthcare, Watertown, MA, 02472, USA\and
Department of Endocrinology and Nutrition, Hospital Universitario Central de Asturias/University of Oviedo, Spain.\and
Inst. Investigación Sanitaria Principado de Asturias (ISPA), Oviedo, Spain.\and
Inst. Universitario  Oncología  Principado de Asturias (IUOPA), Avenida de Roma s/n, 33011, Oviedo, Spain
\email{uo302763@uniovi.es}\and
Bioinspired Intelligence Ltd., Pozuelo de Alarcón, Spain
}

\maketitle              

\begin{abstract}
Genetic programming and its variants, such as grammatical evolution, are widely used in symbolic regression to derive mathematical expressions from multivariate data. In addition to predictive accuracy,  models are appreciated for their potential to provide interpretability, offering explicit equations that relate input variables to outcomes. However, achieving interpretability and plausibility remains challenging, as evolved models may be complex or scientifically inconsistent. In this study, we explore whether large language models (LLMs) can assist in improving the interpretability of symbolic regression models generated by evolutionary computation methods. Building on our previous work on estimating body fat percentage using grammar-based genetic programming, we investigate LLMs as post-processing tools to analyze and rank evolved expressions by interpretability and physiological plausibility. Four symbolic expressions are analyzed by three LLMs over three repeated runs, and the resulting interpretations and rankings are assessed by a panel of three clinicians. Across the three LLMs, comparative model-ranking outputs received more favorable clinician assessments than isolated term-level interpretations. However, the LLMs also produced physiologically and mathematically questionable explanations, indicating that they are better suited to comparative auditing under expert oversight than to autonomous validation.

\end{abstract}

\section{Introduction} 
\label{sec:intro}

Genetic programming (GP) \cite{koza:1992} is a technique used to automatically generate computer programs that can perform specific tasks. One of its most widely used applications is to obtain symbolic regression (SR) \cite{kommenda2015complexity} models from a supervised learning dataset. Unlike many other artificial intelligence (AI) techniques, which yield \textit{black-box} models, SR offers a high degree of interpretability. It generates explicit equations that combine the problem's input variables, allowing us to see how the model produces its predictions and interpret them more directly.

Two properties relate to explainability: interpretability and plausibility. A model is said to be interpretable if the end user can understand the meanings of the terms in a direct manner. However, this is not always the case for SR models. An SR model, i.e., an equation, can have so many terms and be so complex that it may be very difficult to interpret \cite{kommenda2015complexity}.  In this paper, we use interpretability to mean extracting insights from the model. We acknowledge that the literature has no consensus on a formal definition and that it is likely context-dependent.

The main difference between interpretability and explainability, studied in fields such as Explainable Artificial Intelligence (XAI), is the inherent understanding of the model. An interpretable model, such as a mathematical expression obtained with SR, can be understood without input-output pairs to infer its behavior, while XAI studies techniques based on these pairs to explain decisions made by a non-interpretable model. Not all SR models are equally interpretable, though.

On the other hand, a model is considered plausible if, regardless of whether it is interpretable or explainable, it aligns with scientific evidence or proposes physically or medically plausible relationships.  For instance, a model may also contain terms whose units are not dimensionally consistent with the function or variable the model estimates. We may also obtain models that display terms which, although interpretable, reveal relationships that have not been validated by scientific evidence. This last point is a bit different, as the scientific evidence may be incomplete or the reasoning may be biased by what was mentioned or demonstrated in the literature. One domain where this controversy arises is the application of evolutionary algorithms in medicine and healthcare. Finding this evidence in the equations may not be easy for the person responsible for obtaining the model if they lack thorough knowledge of the application area. Similarly, the professional requesting the model may not be familiar with reading and interpreting mathematical equations such as those produced by an SR tool.

We analyze the interpretability and plausibility of models using large language models (LLMs) and physicians. LLMs are deep learning models trained on large volumes of text with the aim of understanding, interpreting, and generating natural language. Thanks to transformer-based architectures \cite{vaswani2017attention}, these models are capable of capturing long-term dependencies and generating coherent text, as well as performing complex natural language processing (NLP) tasks such as summarization, question answering, text classification, and others. Their main contribution is that they can be adapted to specific areas or tasks with minimal additional information based on their prior training with enormous volumes of data (pre-training).

LLMs are also being used to analyze and interpret models obtained using different GP techniques, such as the work in\cite{de2024explainable}, where expressions for detecting hypoglycaemia obtained with grammatical evolution (GE), a version of GP, are analyzed in a straightforward manner. There are other fields where LLMs are applied to obtain interpretation as a post-processing tool for models.
We hypothesized that LLMs can offer interpretability consistent with medical evidence for SR models. 

This paper uses a set of SR models presented in \cite{munoz2025estimation}, where several models using three grammar-guided evolutionary computation methods: GE \cite{Ryan1998}, context-free grammar genetic programming (CFG-GP) \cite{whigham1995grammatically}, and dynamic structured grammatical evolution (DSGE) \cite{lourencco2019structured} were obtained for the estimation of the body fat percentage (BFP). Those GP variants were selected because they enable the helpful configuration of the solution space and generate symbolic expressions that can potentially enhance both predictive performance and model transparency. \cite{munoz2025estimation} compared the mentioned grammar-guided algorithms with the QLattice framework \cite{brolos2021qlattice}, showing that grammatical evolution methods yield competitive and interpretable results. However, the selection of the final model was guided only by the lowest value of the mean square error of the solutions. In this paper, we propose utilizing LLMs to analyze solutions obtained by SR in terms of interpretability and plausibility, and then select the best model accordingly. We analyze solutions from the three methods and compare the alternatives. Additionally, as a main contribution, three physicians study not only the solutions but also the interpretations performed by the LLMs. The results suggest that integrating LLM reasoning into the SR workflow may provide a new path toward interpretable and plausible symbolic models.

The rest of this paper is structured as follows: Section \ref{sec:related_work} presents previous work on the use of LLMs for the interpretability of models. Section \ref{sec:LLMprompting} explains the concepts of LLMs and prompt engineering used in this work. Section \ref{sec:prev_work} collects the resources generated in previous work. Section \ref{sec:methodology} defines the methodology used. Section \ref{sec:results} presents the results obtained with the selected LLMs. Section \ref{sec:conclusions} concludes the paper.

\section{Related Work}
\label{sec:related_work}

The intersection of SR, EC and LLMs has evolved rapidly, shifting from purely syntactic optimization toward hybrid approaches that integrate semantic reasoning. This section structures the existing literature around three fundamental axes: the management of complexity and interpretability in SR, the integration of LLMs as active components within the evolutionary loop, and the emerging role of LLMs for explainable Artificial Intelligence (XAI). We conclude by highlighting the specific gap this work addresses: the use of LLMs as \textit{post-hoc} evaluators to audit the semantic and biomedical plausibility of symbolic models, a critical function that remains largely unexplored.

One of the most persistent challenges in GP applied to either regression or automation is the phenomenon of \textit{bloating}, where syntactic trees grow in size and complexity without offering proportional improvements in accuracy, thereby compromising human interpretability \cite{lacava2021contemporary}. Traditionally, the GP community has addressed this issue through strategies like parsimony pressure and multi-objective optimization. The latter seeks to approximate a Pareto Front that simultaneously optimizes accuracy and syntactic simplicity \cite{lacava2021contemporary, aldeia2025call}. More recent strategies, such as \textit{Age-Fitness Pareto} (AFP), introduce additional objectives like the ``age'' of genetic material to preserve diversity and prevent convergence to complex local optima \cite{schmidt2011age}. However, syntactic complexity is not the sole proxy for interpretability. As argued by \cite{kommenda2015complexity}, nonlinear complexity provides a more robust measure of model interpretability, and modern approaches like LLM-Meta-SR demonstrate that prompt-guided constraints can control bloat while maintaining performance \cite{zhang2026llmmetasrincontextlearningevolving}. Nevertheless, a fundamental challenge persists: even when algorithmic methods effectively control syntactic complexity, the semantic plausibility of the resulting equations within the application domain is not guaranteed. A formula that is mathematically simple and accurate may still be biologically implausible, a dimension that current algorithmic filters cannot evaluate on their own~\cite{aldeia2025call}.

The integration of LLMs has recently transformed the landscape of evolutionary computation, moving beyond their role as interpreters to become active components within the algorithm's lifecycle. This integration manifests in three primary paradigms. First, LLMs act as adaptive evolutionary operators. Systems like EvoPrompt and LEO leverage the contextual rewriting capabilities of LLMs to propose semantically informed variations, replacing or augmenting traditional stochastic operators and improving search efficiency in discrete spaces~\cite{guo2025evoprompt, brahmachary2025large}. Second, LLMs serve as providers of scientific priors: recognizing the inefficiency of search from scratch, frameworks like LLM-SR and CoEvo utilize LLMs to propose equation skeletons based on known principles, which are then refined through numerical optimization, drastically reducing the search space in theory-rich domains~\cite{shojaee2025llmsr, guo2025coevosr}. The most sophisticated approach involves the co-evolution of solutions and concepts, as seen in LASR, where LLMs abstract successful patterns into high-level textual concepts (e.g. logistic growth), creating a dynamic library that feeds back into the evolutionary search, enabling open-ended exploration across multiple representations~\cite{grayeli2024lasr}. Crucially, while these paradigms employ LLMs to generate or enhance equation search, fundamentally optimizing for accuracy or efficiency, they exhibit a critical deficiency regarding the use of LLMs as \textit{post-hoc} evaluators to audit the semantic validity of the solutions.

This semantic validation gap is particularly salient within the field of XAI. Standard model-agnostic methods like LIME \cite{ribeiro2016should} and SHAP \cite{lundberg2017unified} are designed to interpret black-box models by approximating local behavior or assigning feature importance. However, for scientific discovery, these tools present epistemological limitations; they explain \textit{which} variables are important but do not reveal the \textit{functional structure} of the relationship, which is essential for validating physiological plausibility in fields like medicine \cite{rudin2019stop}. SR, by providing structural transparency by design, theoretically obviates the need for such post-hoc approximators. Yet, this advantage only materializes if the expression is interpretable both syntactically and semantically. Here, LLMs represent a opportunity due to their capacity to process natural language and reason over code, positioning them to analyze mathematical expressions as semantic representations of real-world phenomena. Recent surveys confirm the potential of LLMs to bridge complex model outputs with human understanding~\cite{bilal2025llms}, but most applications in XAI focus on explaining black-box models, leaving a gap of work on using LLMs to audit the scientific plausibility of inherently interpretable, white-box models generated by SR.

Motivated by this gap, our work addresses a distinct and critical gap. Unlike the paradigms reviewed above, we do not use the LLM to discover or optimize the equation, a task delegated to GP/GE algorithms, but rather to answer two qualitatively different questions: 
\begin{itemize}
    \item \textit{Can LLMs assist the post-hoc auditing and comparison of symbolic regression models with respect to interpretability and physiological plausibility? }and,
    \item \textit{How well do these assessments align with clinician judgments?}
\end{itemize} We therefore propose a novel perspective where advanced reasoning models operate as an expert common-sense filter, a qualitative auditing layer complementary to quantitative performance metrics. This approach aims to analyze whether SR equations are biologically plausible, an essential step toward building truly trustworthy and adoptable models in knowledge-intensive domains like healthcare, and whether the interpretability and plausibility are detected by an LLM.

\section{Large Language Models and Prompt Engineering}
\label{sec:LLMprompting}

As mentioned above, LLMs can be adapted with little additional information to specific domains or tasks. In addition, unlike traditional LLMs, current LLMs are models whose reasoning is implemented into their normal processing. By combining techniques such as reinforcement learning (RL), ToT and reinforcement learning from human feedback (RLHF), they are able to increase their reasoning performance. Some examples are OpenAI-o1 \cite{jaech2024o1}, and its reduced version, OpenAI-o1-mini, or Deepseek-R1 \cite{guo2025deepseek}.

Nevertheless, there are inherent problems with these models due to the way they are designed and trained. The existence of hallucinations is one of them \cite{zhang2025hallucinations}. Hallucinations are erroneous information generated by the model because it is designed to always respond, even if the provided information is not present in the training data, resulting in false information derived from the actual data. Since LLMs are not designed to generate true or false answers, but to predict the most likely text to a given prompt, the potential for hallucination is inherent in how the models function, and must always be assumed to be potentially present in any text generated by these techniques.

\subsection{Prompt Engineering for Explanation Generation}

Retrieval-Augmented Generation (RAG) grounds the output of an LLM in information retrieved from an external knowledge base, which mitigates hallucinations and the lack of domain-specific knowledge without retraining the model \cite{gao2023rag}; however, it requires a curated corpus from which to retrieve. Fine-tuning, in turn, requires a sufficiently large training dataset. In our setting, neither resource is available. In this context, prompt engineering becomes the most viable strategy to improve the quality of the explanations generated by LLMs.

Prompt engineering is the practice of designing and optimizing instructions or inputs (prompts) to guide the behavior of LLMs. Since these models generate responses based on the context provided in the input text, the way in which the prompt is structured and formulated directly influences the quality, accuracy, and relevance of the output. However, as shown in \cite{wang2024advancedlanguagemodelseliminate}, accurate information is more important than a complex prompt structure for LLMs with reasoning capabilities. Sahoo et al. \cite{sahoo2025systematicsurveypromptengineering} classify the main Prompt engineering techniques in the following taxonomy.

On one hand \textit{Few-Resources prompting} is a prompting technique designed for generating accurate enough outputs without needing extensive data or resources, enabling the base model to perform new tasks. We can distinguish between Zero-Shot and Few-Shot prompting
Zero-Shot prompting  does not need any specific modification of the LLM or examples of execution. The accuracy relies mainly on the creation of the prompt, which must be carefully crafted, following the structure proposed for the actual task. The work in \cite{radford2019ZeroShot} exposes the performance of LLMs such as GPT-3 in different tasks in Zero-Shot settings. Few-Shot prompting \cite{mann2020FewShot} proposes including a few input-output examples in the prompt. With them, the LLM has more data to work on and will have the desired structure for the output since it is already included implicitly in the given examples. Combining the examples with a well-structured prompt, the technique outperforms Zero-Shot. However, the quality of the examples used determines the output of the model, so they must be well constructed. 

On the other hand, \textit{Reasoning prompting} has emerged as a new way of prompting with the creation of the concept of Chain-of-Thought (CoT) \cite{wei2022cot}. CoT prompting aims to compensate for the inability of LLMs to engage in complex reasoning by integrating coherent step-by-step reasoning. This technique allows the model for a deeper understanding of the given prompts. Manual creation of high-quality CoT prompts is time-consuming and suboptimal.\cite{zhang2210autoCot} fixes this problem with the integration of Automatic Chain-of-Thought (Auto-CoT), by using ``Let's think step by step" prompts to generate reasoning chains. This technique minimizes errors and allows for few-shot prompting using CoT without needing the manual creation of high-quality reasoning chains. \cite{long2023tot} and \cite{yao2023tot} propose the use of a Tree-of-Thought (ToT), where instead of working on a linear sequence of thoughts, it allows them to branch out, enabling the use of search algorithms to obtain optimal results. In this work, we use reasoning prompting to optimize the performance of the studied LLMs; in particular, we use Zero-Shot Prompting due to limited resources. Few-shot prompting is strongly dependent on the available data and examples; therefore, it is left as future work.

We propose the following prompt structure and properties to obtain an accurate answer. 
\begin{enumerate}
    \item First, specify the role that the model represents, which can be expressed as an “interpreter of mathematical expressions in the medical field,” followed by a brief description of the assigned task. 
    \item It must be explicitly stated that the output is going to be read by a medical expert, not a mathematician. This specification allows for a better understandability and eliminates any deepening in the mathematical characteristics of the model. 
    \item It is also important to clarify the context of the task. For example, explaining the different variables of the expression, how the expression has been obtained and any other relevant information that may help in narrowing the scope of the generated answers.
    \item Finally, the task and the expression are given. The task must be precise rather than complex, emphasizing how the desired output is structured, as we lack input-output examples that can replace the specifications. Also, the expression has to be understandable by the model, being careful with the symbols used.
\end{enumerate}

\section{Selection of Models}
\label{sec:prev_work}

We analyze four symbolic regression models for estimating total body fat percentage (\texttt{DXDTOPF}) that were generated and validated in a previous study \cite{munoz2025estimation}. These models were selected from a large pool of candidates evolved using two distinct grammar-guided GP variants applied to the NHANES 2017-18 dataset. The SR models analysed in the present study were obtained in that work and are used here as fixed inputs: none of them was generated, retrained or modified for this paper. The four  models are tagged as DSGE-M18, DSGE-M21, CFG-M3, and CFG-M9. DSGE-M21 and DSGE-M18 were obtained using Dynamic Structured Grammatical Evolution (DSGE), while Models CFG-M9 and CFG-M3 obtained through Context-Free Grammar GP (CFG-GP). 

They represent diverse structural characteristics, from relatively compact 6-term expressions (DSGE-M21) to complex formulas with over 20 terms (CFG-M9), while maintaining competitive predictive performance ($R^2 \approx 0.84$-$0.85$ on test data). Table~\ref{tab:models_overview} summarizes the information of the models and their performance metrics. The evolutionary algorithms were applied using Grammatical Modeling Tools (GMT), a specialized Java framework developed by the Absys Research group at Universidad Complutense de Madrid \cite{hidalgo2018identification}. Complete experimental details, including grammars, hyperparameters, and convergence analysis, are provided in \cite{munoz2025estimation}. The models were produced in two stages. First, candidate models were generated through grammar-based GP selecting the best-performing candidates. Then, the selected models were simplified to obtain the final expressions. 

As mentioned above, \cite{munoz2025estimation} use the National Health and Nutrition Examination Survey (NHANES) 2017-2018 dataset \cite{stierman2021national}, a comprehensive cross-sectional study conducted by the U.S. Centers for Disease Control and Prevention (CDC). NHANES provides nationally representative health and nutrition data through interviews, physical examinations, and laboratory tests. The dataset comprises nine anthropometric and demographic features as independent variables and total body fat percentage measured by 
Dual-Energy X-ray Absorptiometry (DXA) as the target variable. Table \ref{tab:variables} presents the variables that are used in the models.

\begin{table}[ht!]
\centering
\caption{NHANES 2017-2018 dataset variables and their descriptions. Target variable shown in bold.}
\label{tab:variables}
\small
\begin{tabular}{llrrrrrrr}
\hline
\textbf{Variable} & \textbf{Description}\\
\hline
RIDAGEYR & Age (years) \\
RIAGENDR & Gender (1=M, 2=F) \\
BMXWT & Weight (kg) \\
BMXHT & Height (cm) \\
BMXLEG & Upper leg length (cm) \\
BMXARML & Upper arm length (cm) \\
BMXARMC & Arm circumference (cm) \\
BMXWAIST & Waist circumference (cm) \\
BMXHIP & Hip circumference (cm) \\
\hline
\textbf{DXDTOPF} & \textbf{Total body fat (\%)} \\
\hline
\end{tabular}
\end{table}

\begin{table}[ht!]
\centering
\caption{Candidate models, the algorithms that generated them, and their performance on the test set; \textit{\# Features} counts the input variables used by each equation, and \textit{Complexity} gives the qualitative level together with the number of operations in the expression.}
\label{tab:models_overview}
\footnotesize
\begin{tabular}{lccccc}
\hline
\textbf{Model} & \textbf{Algorithm} & \textbf{Test $R^2$} & \textbf{Test RMSE} & \textbf{\# Features} & \textbf{Complexity} \\
\hline
DSGE-M21 & DSGE & 0.842 & 3.41 & 6 & Low (14 ops) \\
DSGE-M18 & DSGE & 0.837 & 3.42 & 7 & Low (18 ops) \\
CFG-M9 & CFG-GP & 0.849 & 3.29 & 8 & High (68 ops) \\
CFG-M3 & CFG-GP & 0.844 & 3.34 & 7 & Medium (26 ops) \\
\hline
\end{tabular}
\end{table}

In \cite{munoz2025estimation}, Root Mean Square Error (RMSE) is the primary metric of performance and fitness used to evaluate the prediction. This indicator measures the dispersion and stability of the results, where a lower value reflects a better fit and higher quality of the solution. In the evolutionary process, it is used to guide the selection of the best candidate models by minimizing the difference between predicted and actual values. 

The selection of models prioritized accuracy, choosing mathematical expressions with competitive prediction performance ($R^2 \approx 0.85$).
 The expressions of the models are shown in equations \ref{eq:DSGE-M21} to \ref{eq:CFG-M3}, where $\widehat{\texttt{DXDTOPF}}_{\texttt{Tag}}$ represents the estimated value of Total body fat (\%) given by the model in the $\texttt{Tag}$:

\begin{dmath}
\label{eq:DSGE-M21}
\widehat{\texttt{DXDTOPF}}_{\texttt{DSGE-M21}} = \frac{31 \cdot \texttt{BMXHIP}}{100} + \frac{9 \cdot \texttt{BMXHT} \cdot \texttt{BMXWAIST}}{100000} - \frac{1387 \cdot \texttt{BMXHT}}{130 \cdot \texttt{BMXWAIST}} \quad - \frac{\texttt{BMXWAIST} \cdot \texttt{BMXWT}^2 \cdot \texttt{RIAGENDR}}{540000} + \frac{48 \cdot \texttt{RIAGENDR}}{5} + \frac{\texttt{BMXHT} \cdot \texttt{BMXWAIST}}{\texttt{BMXARML} \cdot \texttt{BMXWT}}
\end{dmath}

\begin{dmath}
\label{eq:DSGE-M18}
\widehat{\texttt{DXDTOPF}}_{\texttt{DSGE-M18}} = - \frac{3 \cdot \texttt{BMXARMC} \cdot \texttt{BMXWAIST} \cdot \texttt{BMXWT}^2 \cdot \texttt{RIAGENDR}}{100000000} - \frac{99 \cdot \texttt{BMXHIP} \cdot \texttt{BMXHT} \cdot \texttt{BMXWT}}{1000 \cdot \texttt{BMXWAIST}^2 \cdot \texttt{RIAGENDR}} \quad + \frac{43 \cdot \texttt{BMXHIP}}{100} + \frac{11 \cdot \texttt{BMXHT}}{2000} + \texttt{RIAGENDR} - \frac{6150 \cdot \texttt{BMXHT} \cdot \texttt{BMXLEG}}{\texttt{BMXARMC} \cdot \texttt{BMXWAIST}^3}
\end{dmath}

\begin{dmath}
\label{eq:CFG-M9}
\widehat{\texttt{DXDTOPF}}_{\texttt{CFG-M9}} =  -\frac{\texttt{BMXHIP}^3 \cdot \texttt{BMXWAIST}}{85000 \cdot \texttt{BMXLEG}^2} - \frac{3 \cdot \texttt{BMXHIP}^2}{31250} + \frac{7 \cdot \texttt{BMXHIP}}{20} - \frac{\texttt{BMXHT}^3}{\texttt{BMXWAIST}^3} + \frac{\texttt{BMXHT}}{\texttt{RIDAGEYR}^2} 
 - \frac{41 \cdot \texttt{BMXHT} \cdot \texttt{BMXWT}}{1000 \cdot \texttt{BMXWAIST}} + \frac{3 \cdot \texttt{BMXWAIST}^2}{31250} + \frac{\texttt{BMXWAIST}}{4100 \cdot \texttt{RIDAGEYR}} + \frac{\texttt{BMXWAIST}}{\texttt{BMXWT}} - \frac{9 \cdot \texttt{BMXWT}}{250} 
 + 7 \cdot \texttt{RIAGENDR} + \frac{5 \cdot \texttt{RIAGENDR}}{\texttt{RIDAGEYR}} + \frac{6 \cdot \texttt{RIDAGEYR}}{625} - \frac{12461}{10000} + \frac{1}{\texttt{RIDAGEYR}} + \frac{\texttt{RIAGENDR}}{\texttt{BMXWT}} - \frac{\texttt{BMXWT}}{\texttt{BMXWAIST}} 
 + \frac{\texttt{BMXLEG} \cdot \texttt{RIAGENDR}}{\texttt{BMXARML}} + \frac{\texttt{RIAGENDR}}{\texttt{BMXARML}} + \frac{\texttt{RIAGENDR}}{\texttt{BMXARMC}}
\end{dmath}

\begin{dmath}
\label{eq:CFG-M3}
\widehat{\texttt{DXDTOPF}}_{\texttt{CFG-M3}}= -\frac{49 \cdot \texttt{BMXARMC}}{500} + \frac{18 \cdot \texttt{BMXARMC}}{\texttt{BMXWT}} + \frac{\texttt{BMXARMC}}{\texttt{BMXWAIST}} - \frac{4 \cdot \texttt{BMXARML}}{\texttt{BMXWT}} - \frac{7 \cdot \texttt{BMXARML}}{\texttt{BMXWAIST}} + \frac{8 \cdot \texttt{BMXHIP}}{25} 
+ \frac{2 \cdot \texttt{BMXHT}}{\texttt{RIDAGEYR}^2} - \frac{7 \cdot \texttt{BMXHT}}{\texttt{BMXWAIST}} - \frac{16 \cdot \texttt{BMXHT}}{5 \cdot \texttt{BMXWAIST} \cdot \texttt{RIAGENDR}} - \frac{\texttt{BMXHT}}{\texttt{BMXWAIST} \cdot \texttt{BMXWT}} + \frac{\texttt{BMXWAIST}}{\texttt{RIDAGEYR}^2} 
 + \frac{4 \cdot \texttt{BMXWAIST}}{\texttt{BMXWT}} + \frac{\texttt{BMXWAIST}}{\texttt{BMXWT} \cdot \texttt{RIAGENDR}} - \frac{23 \cdot \texttt{BMXWT}}{25000000} + \frac{142 \cdot \texttt{RIAGENDR}}{25} + \frac{49}{50} + \frac{49}{50 \cdot \texttt{RIDAGEYR}} 
- \frac{2 \cdot \texttt{BMXWT}}{\texttt{BMXWAIST}} + \frac{\texttt{BMXLEG} \cdot \texttt{BMXWAIST}}{\texttt{BMXHT} \cdot \texttt{BMXWT}} - \frac{3 \cdot \texttt{BMXWT}}{\texttt{BMXHT}} + \frac{\texttt{RIDAGEYR}}{\texttt{BMXHIP}} - \frac{\texttt{BMXHT}}{\texttt{BMXARMC} \cdot \texttt{BMXWT}} + \frac{\texttt{BMXWAIST}}{\texttt{BMXARMC}}
\end{dmath}

\section{Experimental Methodology}
\label{sec:methodology}

\subsection{Selection and Configuration of LLMs}
\label{subsec:llm_framework}

The objective of the LLM selection is to balance the cost of inference with the quality of the output. As shown in \cite{zellinger2025economicevaluationllms}, reasoning models' costs are higher than non-reasoning models. In addition, \cite{Polverini_2025} compares the costs of different models, showing how the cost is not directly related to the accuracy in spatial-reasoning tasks; however, newer and more powerful models are usually more expensive to use. Additionally, access to and control of the largest LLMs is not uniform, as shown in \cite{sathish2024llempowerunderstandingdisparitiescontrol}. We employ a progression of LLMs to assess their capabilities as \textit{post-hoc} interpretability tools. 

We propose the use of an open-source non-reasoning model, an open-source reasoning model (to evaluate an alternative to the options with the greatest disparities) with medium cost, and a cutting-edge Reasoning-enhanced model with high usage costs:

\begin{itemize}
    \item \textbf{Gemma3:} it is a versatile open-source LLM \cite{gemmateam2025gemma3technicalreport} with good reasoning capabilities, but it is not reasoning-oriented, serving as a deterministic baseline compared to reasoning-optimized models. To ensure reproducibility, we used the 27-billion-parameter version with a temperature of 0 (so the model is deterministic) and a fixed seed.
    \item \textbf{DeepSeek-R1:}\cite{Guo_2025} proposes a reasoning-optimized LLM specifically designed for complex logical analysis, allowing us to test if specialized reasoning architectures offer advantages for this task. This model was selected due to its open-source nature and its lower cost of inference compared to newer Reasoning models.
    \item \textbf{GPT-5.1 Thinking:} A reasoning-enhanced LLM from OpenAI, used to examine advancements in structured reasoning and medical knowledge integration.  
\end{itemize}

Furthermore, ensuring structural consistency across different LLMs requires a nuanced approach to generation parameters, particularly the decoding \textit{temperature}. Traditionally, setting a low temperature ($T \approx 0$) has been the standard practice for reducing stochasticity and enforcing determinism in standard LLMs \cite{liu2025temperature}. However, recent literature demonstrates that this parameter adjustment is fundamentally inadequate, and often detrimental, when applied to modern reasoning models.

As highlighted in recent studies such as \cite{looping2025reasoning}, applying greedy decoding or very low temperatures to models that utilize extensive CoT processes frequently forces them into infinite repetition loops or collapses their reasoning boundaries. Because reasoning models inherently require a degree of entropy, defined by the temperature (typically $T \in [0.5, 0.7]$), to successfully navigate complex logic trees without getting stuck, structural consistency and reproducibility cannot be achieved merely by lowering the temperature. Instead, it is pursued through explicit prompt constraints and formatting instructions \cite{schulhoff2024prompt}.

\subsection{LLM-Based Interpretation and Comparison Protocol}
\label{subsec:exp_desing}

The objective is to use the LLMs to interpret and compare the expressions obtained using natural language. This method allows for better comprehension of the model by physicians, avoiding the necessity of knowing how to interpret mathematical expressions. To achieve the objective, we designed a structured protocol consisting of two sequential tasks, the outputs of which will be evaluated and discussed in the results:

\begin{enumerate}
    \item Individual interpretation of each model (applied to Gemma3 and DeepSeek-R1): This task is designed to evaluate the capacity of the LLMs to interpret each model individually, merging the mathematical nature of the models with the possible physiological reasoning behind the different subexpressions. The LLM acts as a cross-domain interpreter, integrating mathematical and clinical knowledge.
    It was assigned the role of a ``\textit{statistician with medical knowledge}" and prompted to:
    \begin{itemize} 
        \item Decompose the mathematical expression into its primary subexpressions.
        \item For each subexpression, provide both a mathematical explanation and a physiological justification based on known body composition principles.
        \item Synthesize these explanations into a coherent natural language summary understandable by clinicians.
    \end{itemize}

    \item Global comparative ranking (applied to all three LLMs): The objective is to set a cross-field comparison, not based solely on performance evaluated with the data, due to the risk of overfitting and the difficulty of interpreting most the complex models. We compare and rank all the proposed models on four criteria derived from clinical interpretability literature:

    \begin{itemize}
        \item \textbf{Interpretability:} Ease of understanding the model's components and logic.
        \item \textbf{Physiological plausibility:} Alignment with established medical knowledge about body fat distribution.
        \item \textbf{Simplicity:} Syntactic complexity and parsimony of the expression.
        \item \textbf{Practical use:} Feasibility for clinical implementation considering measurement requirements and stability.
    \end{itemize}
\end{enumerate}
The output was a rank (1=best to 4=worst) for each criterion, accompanied by brief justifications and an overall recommendation.  

\subsection{Prompt Design} Optimizing the interaction with LLMs requires moving beyond rudimentary instructions towards highly structured prompt engineering. According to the comprehensive taxonomy in \cite{schulhoff2024prompt}, structuring a prompt into discrete components, such as Role, Context, Task, Format, and Constraints, drastically improves performance by reducing computational ambiguity and preventing hallucinations. 

Assigning a specific role is grounded in recent advancements in role-playing language agents. As demonstrated by the \textit{RoleLLM} framework \cite{wang2023rolellm}, explicitly defining a role (e.g., a statistician acting as a bridge between engineers and medics) significantly enhances the model's ability to align its behavioral and conversational style with the target audience. This ensures the generated output strictly adheres to the requested cognitive level, deliberately omitting complex mathematical formulations when addressing users with no statistical background.

The prompt is structured into the following key components:

\begin{itemize}
    \item \textbf{Role:} The model is assigned the role of a ``statistician with both statistical and medical knowledge who interprets results for physicians''.
    
    \item \textbf{Context:} \textit{You work with both AI engineers and physicians. Engineers don't have any medical knowledge, and physicians don't have any mathematical or statistical knowledge, so they need you to work as a bridge and as a selector of models.}
    
    \item \textbf{Task:} The core objective is defined as follows: \textit{Your task is to interpret each of the 4 models (DSGE-M21, DSGE-M18, CFG-M9, CFG-M3) so that a physician with no mathematical knowledge can understand, focusing on each important subexpression, writing it and what it means (you have to consider the sign of the expression and the variables involved, evaluating how the change in each variable affects the outcome), in order of appearance, and ending with a brief summary in natural language. After the interpretations, I want you to do a little comparison of each model in the following points: how easy it is to obtain an interpretation of the expression (Interpretability), how physiologically plausible they are (use your medical knowledge), their simplicity (which is how straightforward the model is), and their practical use in a real situation with new data (I recommend you think in new cases). After the comparison, rank all the 4 models in each of the 4 mentioned points.}

    \item \textbf{Format and Instructions:} To create a highly structured output and avoid overly dense explanations, the model must follow these sequential steps:
    \textit{\begin{enumerate}
        \item Sub-expression $\rightarrow$ Simple medical interpretation (one by one, in order of appearance, but just the most important ones. You have to consider the sign of the expression and the variables involved, evaluating how the change in each variable affects the outcome. Negatives indicate a decrease in the outcome; positives indicate an increase).
        \item Final summary of the model in simple medical language to enhance interpretability.
        \item Repeat the above steps for all 4 models.
        \item Joint comparison of the models according to:
        \begin{itemize}
            \item Ease of interpretation
            \item Physiological plausibility
            \item Simplicity.
            \item Practical use.
        \end{itemize}
        \item Final ranking in each of the aforementioned criteria.
    \end{enumerate}}

    \item \textbf{Constraints:} In order to reduce the entropy of the LLM and standardize the outputs, the prompt includes the following constraints:
   \textit{ \begin{itemize}
        \item Your answers have to be consistent for the same prompt, always give the same answers.
        \item Keep the same format for every model.
        \item Do the structures comparisons and rankings based on each points, for example for interpretability, you have to compare the models, then for physiological plausibility, you have to compare the models and so on.
        \item When you do the comparisons, do the rankings of the 4 models.
        \item Give detailed reasoning of the comparisons and rankings of the 4 models, so you can convince a medic that it is the best decision.
        \item Your reasoning must be understandable by someone without any advanced mathematical knowledge.
        \item Deliver the result in different well structured parts.
    \end{itemize}}
\end{itemize}

\vspace{0.5cm}

The sign-based instruction: ``Negatives indicate a
decrease in the outcome, positives indicate an increase", is a simplifying heuristic for term-level description. Because inputs can occur in multiple nonlinear sub-expressions, the sign of an individual term should not be interpreted as the global marginal effect of a variable on the complete model

We do not create a specific prompt solely for the comparison and ranking of the models for GPT-5.1 Thinking evaluation. This decision was made to preserve the structural integrity of the responses, allowing for a proper benchmarking of the LLMs' outputs. For this reason, GPT-5.1 Thinking has been prompted to perform an individual interpretation of the models, just as Gemma3 and DeepSeek-R1 were prompted to do. However, these individual interpretations are not discussed.

\subsection{Assessment of the LLM Outputs}
\label{subsec:analysis}

The core analysis involves comparing the LLM-generated rankings and explanations across three dimensions. We compare LLM rankings across runs, contrast whether their outputs are correct, and summarize physician ratings descriptively.

\begin{enumerate}
\item \textbf{LLM consistency:} We define it as the agreement between outputs generated by LLMs in their rankings and rationales. This includes intra-LLM consistency (same LLM, different runs) and inter-LLM consistency (different LLMs).
 To assess the intra-LLM consistency, each model was run three times on the same prompt. For the non-reasoning baseline (Gemma3), determinism is enforced through a temperature of 0 and a fixed seed, so repetition serves only to confirm invariance. For the reasoning models (DeepSeek-R1 and GPT-5.1 Thinking), determinism cannot be imposed by lowering the temperature, and consistency is instead pursued through the explicit constraints of the prompt. Repetition, therefore, plays a different role for each type of model: a confirmation for the deterministic baseline and a measurement of run-to-run variability for the reasoning models. To keep the clinical evaluation tractable, each LLM's first run was designated a priori as the reference output and submitted to the panel. The two remaining runs were not evaluated by clinicians but were used to measure ranking consistency; the ranking tables accordingly report three values per cell, one for each run. We evaluate the inter-LLM consistency through a comparison of the rankings generated by the different LLMs.

\item \textbf{Errors and Limitations in LLM Interpretations:} Due to the risk of hallucination in LLM-generated text, all outputs were manually reviewed by two groups of experts, each responsible for a complementary dimension. Physicians assessed the physiological validity of the explanations, contrasting each claim with established clinical knowledge. Computer science researchers assessed the mathematical validity, contrasting each statement about the expression with the equation provided to the model. Claims judged incorrect by either group were recorded as errors. Finally, we analyzed the recorded errors to identify limitations and structural failures of the LLMs for this task.

\item \textbf{Physicians' assessment of the LLM outputs:} 
To ensure the validity of the LLMs' outputs, an evaluation was conducted by three physicians from different hospitals, who are co-authors of the present work. They were given the output of the LLM, the symbolic expression evaluated, a series of statements and they were asked to rate their level of agreement with them on a scale of 1 to 5, depending on the specific task.

The defined statements were the following for each task:

\begin{itemize}
    \item Interpretation of the model:
    \begin{itemize}
        \item The interpretations are physiologically plausible.
        \item The interpretation is comprehensible.
        \item The generated information is relevant.
        \item The explanation provides information that a physician would not have been able to detect at a glance.
        \item Using AI makes it easier for a physician to interpret the model.
    \end{itemize}
    \item Comparison of models:
    \begin{itemize}
        \item The ranking is appropriate.
        \item The reasoning is sound.
        \item The generated information is relevant.
        \item This reasoning provides information that a physician might not have noticed at first glance.
        \item Using AI makes it easier for physicians to compare models
    \end{itemize}
    \item Final Conclusions:
    \begin{itemize}
        \item The conclusion is appropriate.
        \item The generated information is relevant.
        \item The conclusion provides information that a physician would not have been able to detect at a glance.
        \item Using AI makes it easier for physicians to compare models.
    \end{itemize}
    
\end{itemize}
\end{enumerate}

\section{Results}
\label{sec:results}

In this section, we present the results organized around three complementary analyses. Section~\ref{subsec:rankings} examines intra-LLM and inter-LLM ranking consistency, that is, whether each LLM produces the same ranking of the SR models across repeated executions with identical inputs and whether different LLMs produce similar rankings. Section~\ref{subsec:errors} presents a qualitative analysis of the LLMs' outputs, in which selected interpretations and errors are contrasted both mathematically, against the behavior of the SR models, and clinically, against the physicians' comments. Finally, Section~\ref{subsec:phy_eval} reports the physicians' assessment of the two tasks, the individual interpretation of each model and the general comparison between models.

\subsection{Intra-LLM and Inter-LLM Ranking Consistency}
\label{subsec:rankings}
Table~\ref{tab:rankings} shows the rankings generated by the LLMs, where each cell of the table reports the position assigned to an SR model (rows) under a given criterion (columns) in each of the three runs (separated by slashes). Gemma3 ran locally with temperature 0 and produced identical rankings in all three runs. However, DeepSeek-R1 and GPT-5.1 Thinking did not produce consistent rankings across runs. For example, in terms of physiological plausibility, DeepSeek-R1 placed DSGE-M18 second in the first run, first in the second, and third in the third.

\begin{table}[]        
\centering
\begin{tabular}{c | c | c | c | c}
\multicolumn{5}{c}{\textbf{Gemma3}}\\
\textbf{Models} & \textbf{Interpretability} & \textbf{Phy. Plausibility} & \textbf{Simplicity} & \textbf{Pract. Use}  \\
\hline
\textbf{DSGE-M21} & 2/2/2 & 2/2/2 & 2/2/2 & 1/1/1 \\ \hline
\textbf{DSGE-M18} & 3/3/3 & 1/1/1 & 3/3/3 & 2/2/2 \\ \hline
\textbf{CFG-M3}   & 1/1/1 & 4/4/4 & 1/1/1 & 3/3/3 \\ \hline
\textbf{CFG-M9}   & 4/4/4 & 3/3/3 & 4/4/4 & 4/4/4 \\ \hline
\multicolumn{5}{c}{\textbf{DeepSeek-R1}}\\
\textbf{Models} & \textbf{Interpretability} & \textbf{Phy. Plausibility} & \textbf{Simplicity} & \textbf{Pract. Use}  \\
\hline
DSGE-M21  & 1/1/1 & 1/2/1 & 1/1/1 & 1/1/1 \\
\hline
DSGE-M18  & 2/2/3 & 2/1/3 & 2/2/3 & 2/2/3 \\
\hline
CFG-M3   & 3/3/2 & 3/3/2 & 4/3/2 & 3/3/2 \\
\hline
CFG-M9   & 4/4/4 & 4/4/4 & 3/4/4 & 4/4/4 \\
\hline
\multicolumn{5}{c}{\textbf{GPT-5.1 Thinking}}\\
\textbf{Models} & \textbf{Interpretability} & \textbf{Phy. Plausibility} & \textbf{Simplicity} & \textbf{Pract. Use}  \\
\hline
DSGE-M21  & 1/1/1 & 2/2/2 & 1/1/1 & 1/1/1 \\
\hline
DSGE-M18  & 2/2/3 & 4/1/4 & 2/2/2 & 4/2/4 \\
\hline
CFG-M3   & 3/3/2 & 1/3/1 & 4/3/3 & 2/3/2 \\
\hline
CFG-M9   & 4/4/4 & 3/4/3 & 3/4/4 & 3/4/3 \\
\hline
\end{tabular}
\caption{Rankings Generated by the LLMs in different runs.}
    \label{tab:rankings}
\end{table}

Both models consistently placed DSGE-M21 at or near the top and CFG-M9 at or near the bottom, while most of the instability consisted of DSGE-M18 and CFG-M3 exchanging intermediate positions. The most pronounced exception is GPT-5.1 Thinking's assessment of physiological plausibility, where DSGE-M18 moves from last to first and back again across the three runs.

This variability cannot be controlled by the user. Unlike Gemma3, which we ran locally with temperature 0, reasoning models accessed through their APIs do not support temperature settings, so decoding is stochastic by design and run-to-run variation is inherent to how these models are offered.

Determinism under controlled conditions does not automatically extend to other settings. Even with Gemma3 at a temperature 0 on a single device, we observed slight lexical variation between runs, attributable to non-associative floating-point arithmetic in parallel inference \cite{fu2026reproducibilitytokenprobabilitiesexpose, zhang2026deterministicinferencetensorparallel} and to dependence on batch size and tensor-parallel configuration \cite{messina2026introducingbackgroundtemperaturecharacterise, zhang2026deterministicinferencetensorparallel}. Although this variation did not alter the rankings in our setup, these sources of non-determinism depend on hardware and serving configuration, so identical rankings cannot be assumed on different GPU architectures or deployment settings.

This instability has direct implications for clinical use. The panel evaluated only the reference run, so their ratings validate a single sample of each LLM's output rather than its behavior across executions. If a different run had been submitted, the panel would have judged a different ordering, and there is no basis for assuming it would have been rated equally favorably. More importantly, the variability is invisible to the end user: a clinician receiving a single ranking has no indication that a repeated execution might have produced a different one, and each output is presented with the same apparent confidence. The pattern observed in Table~\ref{tab:rankings} sharpens this concern. Rankings were stable for models that were clearly better or clearly worse, and unstable for those occupying intermediate positions, that is, precisely in the cases where the choice between models is least obvious and an auditing tool would be most useful. 

In addition, there is another failure related to rankings, which is the lack of inter-LLM consistency across the rankings. Despite a general agreement on how to rank simplicity and practical use, physiological plausibility and interpretability are ranked differently. The main difference is observed across all the physiological plausibility rankings, where there are inconsistencies in selecting the most plausible model. In terms of interpretability, the only model that disagrees is Gemma3. These failures raise doubts about which LLM to use for the desired task and create an underlying problem, which is the process of evaluating which ranking is the most appropriate.
\subsection{Errors and Limitation in LLM Interpretations}
\label{subsec:errors}


The evaluation conducted by the physicians was crucial to assess physiologically related failures that raise doubts about the viability of running the LLMs without the supervision of a physician. 

The most flagrant issue is the assertion generated several times by Gemma3, which states that for the same anthropometric measures, men have more body fat than women. A plausible explanation of the reason is that Gemma3 does not correctly process some subexpressions, as for example  \begin{equation}
-\frac{\mathrm{BMXWAIST} \times \mathrm{BMXWT}^{2} \times \mathrm{RIAGENDR}}{540000}
\label{eq:dsge-m21-term}
\end{equation}

\noindent from Model DSGE-M21, as it does not take into account the leading negative sign. This behavior recurs across other subexpressions, consistently leading Gemma3 to misinterpret the direction of the effects. 

A different kind of error appears in Model DSGE-M18. Here, Gemma3 reports \begin{equation}
+\frac{11 \times \mathrm{BMXHT}}{2000}
\label{eq:dsge-m18-term}
\end{equation}

\noindent as: \textit{Taller individuals tend to have higher body fat. A 1cm increase in height is estimated to increase body fat by approximately 0.0055\%.} The physicians flagged this statement as physiologically incorrect. The magnitude is also negligible: a 10 cm difference in height translates into 0.055\% of body fat. This does not imply that height is a marginal predictor, as it appears in other terms of the model, but rather that this particular term contributes little to the output, consistent with either bloat or a marginal compensatory term arising from the evolutionary search.

Distinguishing between these two explanations is not possible from the LLM's output, and this points to a broader limitation of term-by-term analysis: interpreting subexpressions in isolation captures local behavior but says nothing about how terms interact globally. A term that appears implausible on its own may be compensating for another, and a term that appears negligible may not be. The same limitation underlies the sign errors reported above. Term-level interpretation is therefore a starting point for expert review rather than a substitute for it.

Another clarifying point concerns the treatment of arm circumference. The physicians noted that this measure does not distinguish between fat and muscle mass. One of them found DeepSeek-R1's treatment of the variable more refined than Gemma3's for model CFG-M3, observing that DeepSeek-R1 attempts to account for why the model uses it with both signs: ``It also interestingly uses arm circumference in both positive and negative contexts, possibly trying to distinguish between muscular and fatty arms". Gemma3, by contrast, restates the term's behavior without addressing its rationale: ``It suggests that a larger arm circumference relative to weight and waist circumference, combined with a smaller waist circumference and taller height, is associated with lower body fat".

The distinction is worth qualifying. DeepSeek-R1's explanation is a plausible hypothesis rather than a verified account, and it attributes intent to a process that has none: the sign pattern emerged from evolutionary search, not from an attempt to separate tissue types. What makes it valuable here is that the hypothesis aligns with the physiological reasoning that one of the clinicians recognizes as sound, which is precisely the kind of judgment the LLM cannot make on its own.

DeepSeek-R1 receives a better evaluation compared to Gemma3, but the main complaint from the physicians is the lack of comprehension when reading the interpretations of DeepSeek-R1. The interpretation generated by the LLM for the model \textit{CFG-M9} is a clear example: ``Initially, hip circumference adds to the body fat percentage, but at very high values, the squared term begins to reduce it, which may not be physiologically intuitive".

\subsection{Physicians' assessment of LLM outputs}
\label{subsec:phy_eval}

\subsubsection*{Comparison of LLMs.}

As discussed in Section \ref{subsec:exp_desing}, both Gemma3 and DeepSeek-R1 produced individual interpretations of the SR models. Comparing them suggests a tension between how readily an interpretation can be understood and how well it holds up to physiological scrutiny: Gemma3's outputs are more direct but frequently assign meaning that does not survive clinical review, whereas DeepSeek-R1's are more tentative and harder to read, but engage more closely with the underlying mechanism. Whether this reflects a genuine trade-off or differences specific to these two models cannot be determined from the present evidence.

Table~\ref{tab:eval_summary} summarises the panel's ratings for both LLMs across the four SR models. Only one dimension separates the two models consistently: DeepSeek-R1's interpretation was rated more physiologically plausible than Gemma3's in three of the four models, and equally plausible in the fourth. Comprehensibility was rated low for both, with medians of 2 or 3 and no separation between them. For the remaining three statements, medians cluster at 3 with substantially overlapping ranges, indicating that neither model was perceived as clearly more useful than the other in terms of relevance, non-obviousness, or ease of interpretation. These figures should be read as descriptive. With three evaluators per cell, the reported medians summarise a small set of ordinal judgments and no statistical testing was performed; differences of one point fall well within what this panel size can produce by chance. The qualitative observations discussed below carry the interpretive weight of this evaluation, with the ratings serving to indicate where the panel converged and where it did not.

\begin{table}[htbp]
    \centering
    \small
    \caption{Physicians' evaluation of LLM interpretations across the four SR models. Values are median (range) over three evaluators on a 1--5 Likert scale, where 1 = strongly disagree and 5 = strongly agree.}
    \label{tab:eval_summary}
    \begin{tabular}{@{}>{\raggedright\arraybackslash}p{4.4cm}
                      >{\centering\arraybackslash}m{1.55cm}
                      >{\centering\arraybackslash}m{1.55cm}
                      >{\centering\arraybackslash}m{1.55cm}
                      >{\centering\arraybackslash}m{1.55cm}@{}}
    \toprule
    \textbf{Statement} & \textbf{DSGE-M18} & \textbf{DSGE-M21} & \textbf{CFG-M3} & \textbf{CFG-M9} \\
    \midrule
    \multicolumn{5}{@{}l}{\textit{The interpretation is physiologically plausible}} \\
    \quad Gemma3      & 2 (1--4) & 2 (2--4) & 3 (2--3) & 3 (2--4) \\
    \quad DeepSeek-R1 & 4 (3--5) & 4 (2--4) & 4 (1--4) & 3 (3--4) \\
    \addlinespace
    \multicolumn{5}{@{}l}{\textit{The interpretation is comprehensible}} \\
    \quad Gemma3      & 2 (2--3) & 2 (2--3) & 2 (1--3) & 2 (2--4) \\
    \quad DeepSeek-R1 & 3 (3--3) & 2 (1--3) & 2 (1--4) & 2 (1--3) \\
    \addlinespace
    \multicolumn{5}{@{}l}{\textit{The generated information is relevant}} \\
    \quad Gemma3      & 3 (2--4) & 3 (3--4) & 3 (2--3) & 3 (3--4) \\
    \quad DeepSeek-R1 & 3 (3--5) & 4 (2--4) & 3 (1--3) & 3 (3--3) \\
    \addlinespace
    \multicolumn{5}{@{}p{\dimexpr\textwidth-2\tabcolsep}@{}}{%
      \textit{The explanation provides information that a physician would not have been able to detect at a glance}} \\
    \quad Gemma3      & 2 (2--5) & 3 (2--4) & 3 (2--3) & 3 (3--3) \\
    \quad DeepSeek-R1 & 3 (3--4) & 3 (2--4) & 3 (1--4) & 3 (3--4) \\
    \addlinespace
    \multicolumn{5}{@{}p{\dimexpr\textwidth-2\tabcolsep}@{}}{%
      \textit{Using AI makes it easier for a physician to interpret the model}} \\
    \quad Gemma3      & 3 (2--4) & 3 (3--4) & 3 (3--4) & 4 (3--4) \\
    \quad DeepSeek-R1 & 3 (3--4) & 3 (3--4) & 3 (2--4) & 3 (3--4) \\
    \bottomrule
    \end{tabular}
\end{table}

\subsubsection*{Individual Interpretation versus General Comparison}

Table~\ref{tab:eval_comparison_summary} shows the panel ratings for the comparative task for the three LLMs, revealing a clear contrast with the individual interpretations reported in Table~\ref{tab:eval_summary}. Medians in the comparative task cluster at 4 across nearly all criteria and all three LLMs, whereas term-level interpretations were predominantly rated 2, leading in interpretability, or 3. Comparative auditing of models was thus received considerably better than isolated interpretation of subexpressions.

In the Conclusion block, in Table~\ref{tab:eval_comparison_summary}, Gemma3 obtained a higher median than the other two models in two of the four statements (\textit{the information is relevant} and \textit{AI eases model comparison}) and the same median in the remaining two, although the ranges overlap in all cases. Gemma3 also showed a wider range in one statement (\textit{AI eases model comparison}: 3--5, versus 3--4 for the other models). The lowest and the highest scores in this block were consistently given by the same evaluators, which suggests that this dispersion reflects individual preferences rather than a systematic property of the output. A possible factor, not tested here, is that its conclusion not only identifies the best model but also proposes alternatives depending on the clinical conditions, which may appeal differently to each evaluator. With only three evaluators, these differences cannot be generalized. In the remaining blocks, the three models had the same median in 11 of the 20 statements. In the other nine, Gemma3 had the lowest median in six (one of them tied with DeepSeek-R1), compared with two for DeepSeek-R1 (one tied with Gemma3) and two for GPT-5.1 Thinking.

\begin{table}[htbp]
    \centering
    \small
    \caption{Physicians' evaluation of the comparative model-ranking outputs produced by the three LLMs. Values are the median (range) over three evaluators on a 1--5 Likert scale, where 1 = strongly disagree and 5 = strongly agree. The first four blocks correspond to the criteria used for ranking; the last reports the overall conclusion.}
    \label{tab:eval_comparison_summary}
    \begin{tabular}{@{}>{\raggedright\arraybackslash}p{5.2cm}
                      >{\centering\arraybackslash}m{1.7cm}
                      >{\centering\arraybackslash}m{1.9cm}
                      >{\centering\arraybackslash}m{1.9cm}@{}}
    \toprule
    \textbf{Statement} & \textbf{Gemma3} & \textbf{DeepSeek-R1} & \textbf{GPT-5.1 Th.} \\
    \midrule
    \multicolumn{4}{@{}l}{\textit{Interpretability}} \\
    \quad The ranking is appropriate            & 4 (4--4) & 4 (3--4) & 4 (4--4) \\
    \quad The reasoning is sound                & 4 (4--4) & 4 (4--4) & 4 (4--4) \\
    \quad The information is relevant           & 4 (3--5) & 4 (3--4) & 4 (3--4) \\
    \quad Not noticed at first glance           & 3 (2--4) & 4 (3--4) & 4 (3--4) \\
    \quad AI eases model comparison             & 4 (3--5) & 3 (3--4) & 4 (3--4) \\
    \addlinespace
    \multicolumn{4}{@{}l}{\textit{Physiological plausibility}} \\
    \quad The ranking is appropriate            & 4 (3--4) & 4 (2--4) & 4 (2--4) \\
    \quad The reasoning is sound                & 4 (3--4) & 4 (2--4) & 4 (3--4) \\
    \quad The information is relevant           & 3 (2--3) & 4 (3--4) & 4 (3--4) \\
    \quad Not noticed at first glance           & 4 (3--4) & 4 (3--4) & 4 (3--4) \\
    \quad AI eases model comparison             & 3 (3--4) & 3 (3--4) & 3 (3--4) \\
    \addlinespace
    \multicolumn{4}{@{}l}{\textit{Simplicity}} \\
    \quad The ranking is appropriate            & 4 (3--4) & 4 (4--4) & 4 (4--4) \\
    \quad The reasoning is sound                & 4 (3--4) & 4 (4--4) & 4 (3--4) \\
    \quad The information is relevant           & 4 (3--4) & 4 (3--4) & 4 (3--4) \\
    \quad Not noticed at first glance           & 3 (2--4) & 3 (2--4) & 4 (3--4) \\
    \quad AI eases model comparison             & 3 (3--4) & 3 (3--4) & 3 (3--4) \\
    \addlinespace
    \multicolumn{4}{@{}l}{\textit{Practical use}} \\
    \quad The ranking is appropriate            & 4 (4--4) & 4 (3--4) & 3 (3--4) \\
    \quad The reasoning is sound                & 4 (4--4) & 4 (3--4) & 3 (3--4) \\
    \quad The information is relevant           & 3 (2--4) & 4 (3--4) & 4 (3--4) \\
    \quad Not noticed at first glance           & 3 (2--4) & 4 (3--4) & 4 (3--4) \\
    \quad AI eases model comparison             & 3 (3--4) & 4 (3--4) & 4 (3--4) \\
    \addlinespace
    \multicolumn{4}{@{}l}{\textit{Conclusion}} \\
    \quad The conclusion is appropriate         & 4 (4--4) & 4 (3--4) & 4 (3--4) \\
    \quad The information is relevant           & 4 (3--4) & 3 (3--4) & 3 (3--4) \\
    \quad Not noticed at first glance           & 4 (2--5) & 4 (3--4) & 4 (3--4) \\
    \quad AI eases model comparison             & 4 (3--5) & 3 (3--4) & 3 (3--4) \\
    \bottomrule
    \end{tabular}
\end{table}

However, all LLMs received similar ratings related to the generated rankings. This is notable because the three models did not produce the same ranking: despite ordering the SR models differently, all three received comparable scores on whether the ranking was appropriate and the reasoning was sound. This convergence reflects a limitation of the evaluation design. The questionnaire asked clinicians to rate each ranking and its reasoning independently, rather than to compare rankings against one another or against a reference ordering. This format was chosen because the clinicians lack the mathematical background needed to rank SR expressions themselves, but it means the instrument captures whether each ranking was judged reasonable, not which of several rankings is better. Within this constraint, the results admit two readings: either the instrument lacks the resolution to separate rankings of differing quality, or the rankings are all defensible, since neither interpretability nor physiological plausibility has a formal definition that would allow models to be strictly ordered: two models can both be plausible without any principled way to establish which is more so. Our data do not allow us to distinguish between these readings.

This also bounds the oversight that physicians can provide. Their ratings can detect reasoning that conflicts with physiological knowledge, but not rankings that are coherent yet incorrect. Establishing a reference ordering would require evaluation formats that do not depend on reading the expressions directly, such as presenting each model's behavior as response curves that can be judged clinically. We leave this for future work.

\section{Conclusions}
\label{sec:conclusions}

This paper addresses whether LLMs can provide interpretability in line with medical evidence for SR models. LLMs do not replace the physician's judgment, but they operate as a useful expert filter: a qualitative auditing layer that, complementing the quantitative performance metrics, helps surface which models are interpretable and physiologically plausible, and which are not.

The clearest finding concerns the difference between the individual interpretation and the general comparison. Across the LLMs, physicians rated the comparative rankings of models more favorably than the interpretations of individual expressions. Since the two tasks were assessed with different questionnaires and a panel of three, this contrast should be read as a descriptive tendency rather than a measured effect.
This does not make individual interpretation without value. By translating each term into clinical language, the LLMs made the expressions accessible to physicians who could not have read them directly, and this is what allowed implausible terms to be identified and discussed. The same interpretations also contained physiologically or mathematically incorrect statements, presented with the same fluency as the correct ones and not flagged by the LLMs themselves; without domain knowledge, such errors are indistinguishable from valid reasoning. This is what justifies treating the LLM as a complementary rather than a substitutive layer: it can organize and articulate the analysis, with comparative ranking the more favorably rated of the two tasks, but a physician must remain in the loop to catch the failures the model cannot detect in itself.

A final observation concerns the consistency of the rankings themselves. The three models produced orderings that differ from one another, and the reasoning models proved inconsistent even with themselves: repeated runs of the same prompt left the top of the ranking stable but shifted the middle positions, whereas the deterministic non-reasoning baseline reproduced an identical ordering every time. Despite this divergence, the clinical assessment of the three models was largely indistinguishable. The most plausible reading is that the physicians anchor their judgment in the soundness of the reasoning rather than in the exact position assigned to each model; with only three evaluators, however, genuine convergence cannot be separated from a lack of resolution in the evaluation instrument itself. This dissociation- variable rankings, homogeneous reception- is a finding to be confirmed rather than a settled result.

The results do not support using LLMs as autonomous arbiters of physiological plausibility for symbolic regression models. They do, however, point to two narrower roles that appear feasible. First, LLMs can organize comparative assessments of candidate symbolic models, but require mathematical verification and clinician oversight. Second, by translating mathematical expressions into natural language, LLMs may act as interpreters between the models and the clinicians: in our evaluation, this translation made the expressions accessible to physicians without mathematical training and allowed them to identify relationships that conflict with physiological knowledge. In this role, the LLM does not judge plausibility itself, which remains the physician's task, but makes that judgment possible for expressions that would otherwise be opaque to clinical review. Both roles rest on a small panel and a limited set of models, so they should be read as indications of viability rather than established uses.

These conclusions open several lines of work. The clinical panel should be widened to a larger and more diverse set of evaluators, which would both strengthen the validation and clarify whether the dissociation just described survives a more sensitive instrument. The physiological errors detected by the clinicians motivate equipping the LLM with precomputed analytical information about each model, the direction and magnitude of each variable's effect, so that its plausibility judgments rest on the model's actual behavior rather than on its surface form. Finally, the run-to-run variability of the reasoning models invites a more systematic study of how to enforce consistency in tasks of this kind. Together, these directions point toward LLMs as an increasingly reliable, though always complementary, component of an explainable symbolic regression workflow in medicine.

\section*{Acknowledgments}
This work is funded by the Spanish Ministry of Science, Innovation, and Universities and the Agencia Espa\~nola de Investigaci\'on, through grants PDC2025-165077-I00, PID2024-158129OB-I00 MICIU/AEI/10.13039/501100011033 and by FEDER, UE.

\subsubsection*{Generative AI Statement.}
Claude Sonnet 5.5 was used for proofreading 
and improving the English language of the manuscript. The authors take full responsibility for the content of the publication.

\bibliographystyle{splncs04}

\bibliography{2025-10-02_bioinspired,bioinspired,biblio_extra}

@article{de2024explainable,
  title={Explainable hypoglycemia prediction models through dynamic structured grammatical evolution},
  author={De La Cruz, Marina and Garnica, Oscar and Cervigon, Carlos and Velasco, Jose Manuel and Hidalgo, J Ignacio},
  journal={Scientific Reports},
  volume={14},
  number={1},
  pages={12591},
  year={2024},
  publisher={Nature Publishing Group UK London}
}

@inproceedings{munoz2025estimation,
  title={Estimation of total body fat using symbolic regression and evolutionary algorithms},
  author={Mu{\~n}oz, Jose-Manuel and Mor{\'o}n-Garc{\'\i}a, Odin and Costilla-Reyes, Omar and Hidalgo, J Ignacio},
  booktitle={International Conference on the Applications of Evolutionary Computation (Part of EvoStar)},
  pages={507--521},
  year={2025},
  organization={Springer}
}

@inproceedings{lourencco2019structured,
  title={Structured grammatical evolution for glucose prediction in diabetic patients},
  author={Louren{\c{c}}o, Nuno and Colmenar, J Manuel and Hidalgo, J Ignacio and Garnica, {\'O}scar},
  booktitle={Proceedings of the Genetic and Evolutionary Computation Conference},
  pages={1250--1257},
  year={2019},
  organization={ACM}
}

@incollection{hidalgo2018identification,
  title={Identification of Models for Glucose Blood Values in Diabetics by Grammatical Evolution},
  author={Hidalgo, J Ignacio and Colmenar, J Manuel and Velasco, J Manuel and Kronberger, Gabriel and Winkler, Stephan M and Garnica, Oscar and Lanchares, Juan},
  booktitle={Handbook of Grammatical Evolution},
  pages={367--393},
  year={2018},
  publisher={Springer}
}

@inproceedings{kommenda2015complexity,
  title={Complexity Measures for Multi-objective Symbolic Regression},
  author={Kommenda, Michael and Beham, Andreas and Affenzeller, Michael and Kronberger, Gabriel},
  booktitle={International Conference on Computer Aided Systems Theory},
  pages={409--416},
  year={2015},
  organization={Springer}
}

@BOOK{koza:1992,
      author = "John R. Koza",
      title = "Genetic Programming: On the Programming of Computers by Means of Natural Selection",
      publisher = "The MIT Press",
      year = "1992"
  }

@incollection{Ryan1998,
	Affiliation = {University of Limerick Dept. Of Computer Science and Information Systems Ireland Ireland},
	Author = {Ryan, Conor and Collins, JJ and Neill, Michael},
	Booktitle = {Genetic Programming},
	Editor = {Banzhaf, Wolfgang and Poli, Riccardo and Schoenauer, Marc and Fogarty, Terence},
	Isbn = {978-3-540-64360-9},
	Pages = {83-96},
	Publisher = {Springer Berlin / Heidelberg},
	Series = {Lecture Notes in Computer Science},
	Title = {Grammatical evolution: Evolving programs for an arbitrary language},
	Volume = {1391},
	Year = {1998}}

@misc{zhang2026llmmetasrincontextlearningevolving,
      title={LLM-Meta-SR: In-Context Learning for Evolving Selection Operators in Symbolic Regression}, 
      author={Hengzhe Zhang and Qi Chen and Bing Xue and Wolfgang Banzhaf and Mengjie Zhang},
      year={2026},
      eprint={2505.18602},
      archivePrefix={arXiv},
      primaryClass={cs.NE},
      url={https://arxiv.org/abs/2505.18602}, 
}

@misc{zhang2026deterministicinferencetensorparallel,
      title={Deterministic Inference across Tensor Parallel Sizes That Eliminates Training-Inference Mismatch}, 
      author={Ziyang Zhang and Xinheng Ding and Jiayi Yuan and Rixin Liu and Huizi Mao and Jiarong Xing and Zirui Liu},
      year={2026},
      eprint={2511.17826},
      archivePrefix={arXiv},
      primaryClass={cs.LG},
      url={https://arxiv.org/abs/2511.17826}, 
}

@misc{fu2026reproducibilitytokenprobabilitiesexpose,
      title={Beyond Reproducibility: Token Probabilities Expose Large Language Model Nondeterminism}, 
      author={Tairan Fu and Gonzalo Martínez and Javier Conde and Carlos Arriaga and Pedro Reviriego and Xiuyuan Qi and Shanshan Liu},
      year={2026},
      eprint={2601.06118},
      archivePrefix={arXiv},
      primaryClass={cs.AI},
      url={https://arxiv.org/abs/2601.06118}, 
}

@misc{messina2026introducingbackgroundtemperaturecharacterise,
      title={Introducing Background Temperature to Characterise Hidden Randomness in Large Language Models}, 
      author={Alberto Messina and Stefano Scotta},
      year={2026},
      eprint={2604.22411},
      archivePrefix={arXiv},
      primaryClass={cs.AI},
      url={https://arxiv.org/abs/2604.22411}, 
}

@misc{guo2025coevosr,
      title={CoEvo: Continual Evolution of Symbolic Solutions Using Large Language Models}, 
      author={Ping Guo and Qingfu Zhang and Xi Lin},
      year={2025},
      eprint={2412.18890},
      archivePrefix={arXiv},
      primaryClass={cs.AI},
      url={https://arxiv.org/abs/2412.18890}, 
}

@inproceedings{grayeli2024lasr,
author = {Grayeli, Arya and Sehgal, Atharva and Costilla-Reyes, Omar and Cranmer, Miles and Chaudhuri, Swarat},
title = {Symbolic regression with a learned concept library},
year = {2024},
isbn = {9798331314385},
publisher = {Curran Associates Inc.},
address = {Red Hook, NY, USA},
booktitle = {Proceedings of the 38th International Conference on Neural Information Processing Systems},
articleno = {1419},
numpages = {32},
location = {Vancouver, BC, Canada},
series = {NIPS '24}
}

@misc{shojaee2025llmsr,
      title={LLM-SR: Scientific Equation Discovery via Programming with Large Language Models}, 
      author={Parshin Shojaee and Kazem Meidani and Shashank Gupta and Amir Barati Farimani and Chandan K Reddy},
      year={2025},
      eprint={2404.18400},
      archivePrefix={arXiv},
      primaryClass={cs.LG},
      url={https://arxiv.org/abs/2404.18400}, 
}

@article{gao2023rag,
        title={Retrieval-Augmented Generation for Large Language Models: A Survey},
        author={Yun-Fan Gao and Yun Xiong and Xin-Yu Gao and Kang-Xiang Jia and Jin Pan and Yuxi Bi and Yi Dai and Jia-Wei Sun and Qian-Yu Guo and Meng Wang and Hao-Fen Wang},
        journal={ArXiv},year={2023},
        volume={abs/2312.10997},
        doi={10.48550/arxiv.2312.10997}
        }

@misc{brolos2021qlattice,
      title={An Approach to Symbolic Regression Using Feyn}, 
      author={Kevin René Broløs and Meera Vieira Machado and Chris Cave and Jaan Kasak and Valdemar Stentoft-Hansen and Victor Galindo Batanero and Tom Jelen and Casper Wilstrup},
      year={2021},
      eprint={2104.05417},
      archivePrefix={arXiv},
      primaryClass={cs.LG},
      url={https://arxiv.org/abs/2104.05417}, 
}

@article{stierman2021national,
  title={National Health and Nutrition Examination Survey 2017-March 2020 prepandemic data files-development of files and prevalence estimates for selected health outcomes},
  author={Stierman, Bryan and Afful, Joseph and Carroll, Margaret D and Chen, Te-Ching and Davy, Orlando and Fink, Steven and Fryar, Cheryl D and Gu, Qiuping and Hales, Craig M and Hughes, Jeffery P and others},
  journal={National health statistics reports},
  number={158},
  pages={10--15620},
  year={2021}
}

@inproceedings{whigham1995grammatically,
  title={Grammatically-based genetic programming},
  author={Whigham, Peter A and others},
  booktitle={Proceedings of the workshop on genetic programming: from theory to real-world applications},
  volume={16},
  number={3},
  pages={33--41},
  year={1995},
  organization={Citeseer}
}

@article{radford2019ZeroShot,
  title={Language models are unsupervised multitask learners},
  author={Radford, Alec and Wu, Jeffrey and Child, Rewon and Luan, David and Amodei, Dario and Sutskever, Ilya and others},
  journal={OpenAI blog},
  volume={1},
  number={8},
  pages={9},
  year={2019}
}

@article{mann2020FewShot,
  title={Language models are few-shot learners},
  author={Mann, Ben and Ryder, Nick and Subbiah, Melanie and Kaplan, J and Dhariwal, P and Neelakantan, A and Shyam, P and Sastry, G and Askell, A and Agarwal, S and others},
  journal={arXiv preprint arXiv:2005.14165},
  volume={1},
  number={3},
  pages={3},
  year={2020}
}

@article{wei2022cot,
  title={Chain-of-thought prompting elicits reasoning in large language models},
  author={Wei, Jason and Wang, Xuezhi and Schuurmans, Dale and Bosma, Maarten and Xia, Fei and Chi, Ed and Le, Quoc V and Zhou, Denny and others},
  journal={Advances in neural information processing systems},
  volume={35},
  pages={24824--24837},
  year={2022}
}

@article{zhang2210autoCot,
  title={Automatic chain of thought prompting in large language models. arXiv 2022},
  author={Zhang, Zhuosheng and Zhang, Aston and Li, Mu and Smola, Alex},
  journal={arXiv preprint arXiv:2210.03493},
year={2022}
}

@article{jaech2024o1,
  title={Openai o1 system card},
  author={Jaech, Aaron and Kalai, Adam and Lerer, Adam and Richardson, Adam and El-Kishky, Ahmed and Low, Aiden and Helyar, Alec and Madry, Aleksander and Beutel, Alex and Carney, Alex and others},
  journal={arXiv preprint arXiv:2412.16720},
  year={2024}
}

@article{guo2025deepseek,
  title={Deepseek-r1: Incentivizing reasoning capability in llms via reinforcement learning},
  author={Guo, Daya and Yang, Dejian and Zhang, Haowei and Song, Junxiao and Zhang, Ruoyu and Xu, Runxin and Zhu, Qihao and Ma, Shirong and Wang, Peiyi and Bi, Xiao and others},
  journal={arXiv preprint arXiv:2501.12948},
  year={2025}
}

@misc{wang2024advancedlanguagemodelseliminate,
      title={Do Advanced Language Models Eliminate the Need for Prompt Engineering in Software Engineering?}, 
      author={Guoqing Wang and Zeyu Sun and Zhihao Gong and Sixiang Ye and Yizhou Chen and Yifan Zhao and Qingyuan Liang and Dan Hao},
      year={2024},
      eprint={2411.02093},
      archivePrefix={arXiv},
      primaryClass={cs.SE},
      url={https://arxiv.org/abs/2411.02093}, 
}

@misc{sahoo2025systematicsurveypromptengineering,
      title={A Systematic Survey of Prompt Engineering in Large Language Models: Techniques and Applications}, 
      author={Pranab Sahoo and Ayush Kumar Singh and Sriparna Saha and Vinija Jain and Samrat Mondal and Aman Chadha},
      year={2025},
      eprint={2402.07927},
      archivePrefix={arXiv},
      primaryClass={cs.AI},
      url={https://arxiv.org/abs/2402.07927}, 
}

@article{vaswani2017attention,
  title={Attention is all you need},
  author={Vaswani, Ashish and Shazeer, Noam and Parmar, Niki and Uszkoreit, Jakob and Jones, Llion and Gomez, Aidan N and Kaiser, {\L}ukasz and Polosukhin, Illia},
  journal={Advances in neural information processing systems},
  volume={30},
  year={2017}
}

@article{long2023tot,
  title={Large language model guided tree-of-thought},
  author={Long, Jieyi},
  journal={arXiv preprint arXiv:2305.08291},
  year={2023}
}

@article{yao2023tot,
  title={Tree of thoughts: Deliberate problem solving with large language models},
  author={Yao, Shunyu and Yu, Dian and Zhao, Jeffrey and Shafran, Izhak and Griffiths, Tom and Cao, Yuan and Narasimhan, Karthik},
  journal={Advances in neural information processing systems},
  volume={36},
  pages={11809--11822},
  year={2023}
}

@article{zhang2025hallucinations,
  title={Siren’s Song in the AI Ocean: A Survey on Hallucination in Large Language Models},
  author={Zhang, Yue and Li, Yafu and Cui, Leyang and Cai, Deng and Liu, Lemao and Fu, Tingchen and Huang, Xinting and Zhao, Enbo and Zhang, Yu and Chen, Yulong and others},
  journal={Computational Linguistics},
  pages={1--46},
  year={2025},
  publisher={MIT Press 255 Main Street, 9th Floor, Cambridge, Massachusetts 02142, USA~…}
}

@article{lacava2021contemporary,
  title={Contemporary symbolic regression methods and their relative performance},
  author={La Cava, William and Burlacu, Bogdan and Virgolin, Marco and Kommenda, Michael and Orzechowski, Patryk and de Fran{\c{c}}a, Fabr{\'\i}cio Olivetti and Jin, Ying and Moore, Jason H},
  journal={Advances in neural information processing systems},
  volume={2021},
  number={DB1},
  pages={1},
  year={2021}
}

@inproceedings{aldeia2025call,
  title={Call for Action: towards the next generation of symbolic regression benchmark},
  author={Imai Aldeia, Guilherme Seidyo and Zhang, Hengzhe and Bomarito, Geoffrey and Cranmer, Miles and Fonseca, Alcides and Burlacu, Bogdan and La Cava, William G and de Fran{\c{c}}a, Fabr{\'\i}cio Olivetti},
  booktitle={Proceedings of the Genetic and Evolutionary Computation Conference Companion},
  pages={2529--2538},
  year={2025}
}

@Inbook{schmidt2011age,
author="Schmidt, Michael
and Lipson, Hod",
editor="Riolo, Rick
and McConaghy, Trent
and Vladislavleva, Ekaterina",
title="Age-Fitness Pareto Optimization",
bookTitle="Genetic Programming Theory and Practice VIII",
year="2011",
publisher="Springer New York",
address="New York, NY",
pages="129--146",
isbn="978-1-4419-7747-2",
doi="10.1007/978-1-4419-7747-2_8",
url="https://doi.org/10.1007/978-1-4419-7747-2_8"
}

@article{guo2025evoprompt,
  title={EvoPrompt: Connecting LLMs with Evolutionary Algorithms Yields Powerful Prompt Optimizers},
  author={Guo, Qingyan and Wang, Rui and Guo, Junliang and Li, Bei and Song, Kaitao and Tan, Xu and Liu, Guoqing and Bian, Jiang and Yang, Yujiu},
  journal={arXiv preprint arXiv:2309.08532},
  year={2025}
}

@article{rudin2019stop,
  title={Stop explaining black box machine learning models for high stakes decisions and use interpretable models instead},
  author={Rudin, Cynthia},
  journal={Nature machine intelligence},
  volume={1},
  number={5},
  pages={206--215},
  year={2019},
  publisher={Nature Publishing Group UK London}
}

@article{bilal2025llms,
  title={Llms for explainable ai: A comprehensive survey},
  author={Bilal, Ahsan and Ebert, David and Lin, Beiyu},
  journal={arXiv preprint arXiv:2504.00125},
  year={2025}
}

@article{brahmachary2025large,
  title={Large language model-based evolutionary optimizer: Reasoning with elitism},
  author={Brahmachary, Shuvayan and Joshi, Subodh M and Panda, Aniruddha and Koneripalli, Kaushik and Sagotra, Arun Kumar and Patel, Harshil and Sharma, Ankush and Jagtap, Ameya D and Kalyanaraman, Kaushic},
  journal={Neurocomputing},
  volume={622},
  pages={129272},
  year={2025},
  publisher={Elsevier}
}

@inproceedings{ribeiro2016should,
  title={" Why should i trust you?" Explaining the predictions of any classifier},
  author={Ribeiro, Marco Tulio and Singh, Sameer and Guestrin, Carlos},
  booktitle={Proceedings of the 22nd ACM SIGKDD international conference on knowledge discovery and data mining},
  pages={1135--1144},
  year={2016}
}

@article{lundberg2017unified,
  title={A unified approach to interpreting model predictions},
  author={Lundberg, Scott M and Lee, Su-In},
  journal={Advances in neural information processing systems},
  volume={30},
  year={2017}
}

@article{Polverini_2025,
   title={Multimodal large language models and physics visual tasks: comparative analysis of performance and costs},
   volume={46},
   ISSN={1361-6404},
   url={http://dx.doi.org/10.1088/1361-6404/ae03f8},
   DOI={10.1088/1361-6404/ae03f8},
   number={5},
   journal={European Journal of Physics},
   publisher={IOP Publishing},
   author={Polverini, Giulia and Gregorcic, Bor},
   year={2025},
   month=sep, pages={055708} }

@article{Guo_2025,
   title={DeepSeek-R1 incentivizes reasoning in LLMs through reinforcement learning},
   volume={645},
   ISSN={1476-4687},
   url={http://dx.doi.org/10.1038/s41586-025-09422-z},
   DOI={10.1038/s41586-025-09422-z},
   number={8081},
   journal={Nature},
   publisher={Springer Science and Business Media LLC},
   author={Guo, Daya and Yang, Dejian and Zhang, Haowei and Song, Junxiao and Wang, Peiyi and Zhu, Qihao and Xu, Runxin and Zhang, Ruoyu and Ma, Shirong and Bi, Xiao and Zhang, Xiaokang and Yu, Xingkai and Wu, Yu and Wu, Z. F. and Gou, Zhibin and Shao, Zhihong and Li, Zhuoshu and Gao, Ziyi and Liu, Aixin and Xue, Bing and Wang, Bingxuan and Wu, Bochao and Feng, Bei and Lu, Chengda and Zhao, Chenggang and Deng, Chengqi and Ruan, Chong and Dai, Damai and Chen, Deli and Ji, Dongjie and Li, Erhang and Lin, Fangyun and Dai, Fucong and Luo, Fuli and Hao, Guangbo and Chen, Guanting and Li, Guowei and Zhang, H. and Xu, Hanwei and Ding, Honghui and Gao, Huazuo and Qu, Hui and Li, Hui and Guo, Jianzhong and Li, Jiashi and Chen, Jingchang and Yuan, Jingyang and Tu, Jinhao and Qiu, Junjie and Li, Junlong and Cai, J. L. and Ni, Jiaqi and Liang, Jian and Chen, Jin and Dong, Kai and Hu, Kai and You, Kaichao and Gao, Kaige and Guan, Kang and Huang, Kexin and Yu, Kuai and Wang, Lean and Zhang, Lecong and Zhao, Liang and Wang, Litong and Zhang, Liyue and Xu, Lei and Xia, Leyi and Zhang, Mingchuan and Zhang, Minghua and Tang, Minghui and Zhou, Mingxu and Li, Meng and Wang, Miaojun and Li, Mingming and Tian, Ning and Huang, Panpan and Zhang, Peng and Wang, Qiancheng and Chen, Qinyu and Du, Qiushi and Ge, Ruiqi and Zhang, Ruisong and Pan, Ruizhe and Wang, Runji and Chen, R. J. and Jin, R. L. and Chen, Ruyi and Lu, Shanghao and Zhou, Shangyan and Chen, Shanhuang and Ye, Shengfeng and Wang, Shiyu and Yu, Shuiping and Zhou, Shunfeng and Pan, Shuting and Li, S. S. and Zhou, Shuang and Wu, Shaoqing and Yun, Tao and Pei, Tian and Sun, Tianyu and Wang, T. and Zeng, Wangding and Liu, Wen and Liang, Wenfeng and Gao, Wenjun and Yu, Wenqin and Zhang, Wentao and Xiao, W. L. and An, Wei and Liu, Xiaodong and Wang, Xiaohan and Chen, Xiaokang and Nie, Xiaotao and Cheng, Xin and Liu, Xin and Xie, Xin and Liu, Xingchao and Yang, Xinyu and Li, Xinyuan and Su, Xuecheng and Lin, Xuheng and Li, X. Q. and Jin, Xiangyue and Shen, Xiaojin and Chen, Xiaosha and Sun, Xiaowen and Wang, Xiaoxiang and Song, Xinnan and Zhou, Xinyi and Wang, Xianzu and Shan, Xinxia and Li, Y. K. and Wang, Y. Q. and Wei, Y. X. and Zhang, Yang and Xu, Yanhong and Li, Yao and Zhao, Yao and Sun, Yaofeng and Wang, Yaohui and Yu, Yi and Zhang, Yichao and Shi, Yifan and Xiong, Yiliang and He, Ying and Piao, Yishi and Wang, Yisong and Tan, Yixuan and Ma, Yiyang and Liu, Yiyuan and Guo, Yongqiang and Ou, Yuan and Wang, Yuduan and Gong, Yue and Zou, Yuheng and He, Yujia and Xiong, Yunfan and Luo, Yuxiang and You, Yuxiang and Liu, Yuxuan and Zhou, Yuyang and Zhu, Y. X. and Huang, Yanping and Li, Yaohui and Zheng, Yi and Zhu, Yuchen and Ma, Yunxian and Tang, Ying and Zha, Yukun and Yan, Yuting and Ren, Z. Z. and Ren, Zehui and Sha, Zhangli and Fu, Zhe and Xu, Zhean and Xie, Zhenda and Zhang, Zhengyan and Hao, Zhewen and Ma, Zhicheng and Yan, Zhigang and Wu, Zhiyu and Gu, Zihui and Zhu, Zijia and Liu, Zijun and Li, Zilin and Xie, Ziwei and Song, Ziyang and Pan, Zizheng and Huang, Zhen and Xu, Zhipeng and Zhang, Zhongyu and Zhang, Zhen},
   year={2025},
   month=sep, pages={633–638} }

@misc{zellinger2025economicevaluationllms,
      title={Economic Evaluation of LLMs}, 
      author={Michael J. Zellinger and Matt Thomson},
      year={2025},
      eprint={2507.03834},
      archivePrefix={arXiv},
      primaryClass={cs.AI},
      url={https://arxiv.org/abs/2507.03834}, 
}

@misc{sathish2024llempowerunderstandingdisparitiescontrol,
      title={LLeMpower: Understanding Disparities in the Control and Access of Large Language Models}, 
      author={Vishwas Sathish and Hannah Lin and Aditya K Kamath and Anish Nyayachavadi},
      year={2024},
      eprint={2404.09356},
      archivePrefix={arXiv},
      primaryClass={cs.CY},
      url={https://arxiv.org/abs/2404.09356}, 
}

@misc{gemmateam2025gemma3technicalreport,
      title={Gemma 3 Technical Report}, 
      author={Gemma Team and Aishwarya Kamath and Johan Ferret and Shreya Pathak and Nino Vieillard and Ramona Merhej and Sarah Perrin and Tatiana Matejovicova and Alexandre Ramé and Morgane Rivière and Louis Rouillard and Thomas Mesnard and Geoffrey Cideron and Jean-bastien Grill and Sabela Ramos and Edouard Yvinec and Michelle Casbon and Etienne Pot and Ivo Penchev and Gaël Liu and Francesco Visin and Kathleen Kenealy and Lucas Beyer and Xiaohai Zhai and Anton Tsitsulin and Robert Busa-Fekete and Alex Feng and Noveen Sachdeva and Benjamin Coleman and Yi Gao and Basil Mustafa and Iain Barr and Emilio Parisotto and David Tian and Matan Eyal and Colin Cherry and Jan-Thorsten Peter and Danila Sinopalnikov and Surya Bhupatiraju and Rishabh Agarwal and Mehran Kazemi and Dan Malkin and Ravin Kumar and David Vilar and Idan Brusilovsky and Jiaming Luo and Andreas Steiner and Abe Friesen and Abhanshu Sharma and Abheesht Sharma and Adi Mayrav Gilady and Adrian Goedeckemeyer and Alaa Saade and Alex Feng and Alexander Kolesnikov and Alexei Bendebury and Alvin Abdagic and Amit Vadi and András György and André Susano Pinto and Anil Das and Ankur Bapna and Antoine Miech and Antoine Yang and Antonia Paterson and Ashish Shenoy and Ayan Chakrabarti and Bilal Piot and Bo Wu and Bobak Shahriari and Bryce Petrini and Charlie Chen and Charline Le Lan and Christopher A. Choquette-Choo and CJ Carey and Cormac Brick and Daniel Deutsch and Danielle Eisenbud and Dee Cattle and Derek Cheng and Dimitris Paparas and Divyashree Shivakumar Sreepathihalli and Doug Reid and Dustin Tran and Dustin Zelle and Eric Noland and Erwin Huizenga and Eugene Kharitonov and Frederick Liu and Gagik Amirkhanyan and Glenn Cameron and Hadi Hashemi and Hanna Klimczak-Plucińska and Harman Singh and Harsh Mehta and Harshal Tushar Lehri and Hussein Hazimeh and Ian Ballantyne and Idan Szpektor and Ivan Nardini and Jean Pouget-Abadie and Jetha Chan and Joe Stanton and John Wieting and Jonathan Lai and Jordi Orbay and Joseph Fernandez and Josh Newlan and Ju-yeong Ji and Jyotinder Singh and Kat Black and Kathy Yu and Kevin Hui and Kiran Vodrahalli and Klaus Greff and Linhai Qiu and Marcella Valentine and Marina Coelho and Marvin Ritter and Matt Hoffman and Matthew Watson and Mayank Chaturvedi and Michael Moynihan and Min Ma and Nabila Babar and Natasha Noy and Nathan Byrd and Nick Roy and Nikola Momchev and Nilay Chauhan and Noveen Sachdeva and Oskar Bunyan and Pankil Botarda and Paul Caron and Paul Kishan Rubenstein and Phil Culliton and Philipp Schmid and Pier Giuseppe Sessa and Pingmei Xu and Piotr Stanczyk and Pouya Tafti and Rakesh Shivanna and Renjie Wu and Renke Pan and Reza Rokni and Rob Willoughby and Rohith Vallu and Ryan Mullins and Sammy Jerome and Sara Smoot and Sertan Girgin and Shariq Iqbal and Shashir Reddy and Shruti Sheth and Siim Põder and Sijal Bhatnagar and Sindhu Raghuram Panyam and Sivan Eiger and Susan Zhang and Tianqi Liu and Trevor Yacovone and Tyler Liechty and Uday Kalra and Utku Evci and Vedant Misra and Vincent Roseberry and Vlad Feinberg and Vlad Kolesnikov and Woohyun Han and Woosuk Kwon and Xi Chen and Yinlam Chow and Yuvein Zhu and Zichuan Wei and Zoltan Egyed and Victor Cotruta and Minh Giang and Phoebe Kirk and Anand Rao and Kat Black and Nabila Babar and Jessica Lo and Erica Moreira and Luiz Gustavo Martins and Omar Sanseviero and Lucas Gonzalez and Zach Gleicher and Tris Warkentin and Vahab Mirrokni and Evan Senter and Eli Collins and Joelle Barral and Zoubin Ghahramani and Raia Hadsell and Yossi Matias and D. Sculley and Slav Petrov and Noah Fiedel and Noam Shazeer and Oriol Vinyals and Jeff Dean and Demis Hassabis and Koray Kavukcuoglu and Clement Farabet and Elena Buchatskaya and Jean-Baptiste Alayrac and Rohan Anil and Dmitry and Lepikhin and Sebastian Borgeaud and Olivier Bachem and Armand Joulin and Alek Andreev and Cassidy Hardin and Robert Dadashi and Léonard Hussenot},
      year={2025},
      eprint={2503.19786},
      archivePrefix={arXiv},
      primaryClass={cs.CL},
      url={https://arxiv.org/abs/2503.19786}, 
}

@misc{schulhoff2024prompt,
      title={The Prompt Report: A Systematic Survey of Prompt Engineering Techniques}, 
      author={Sander Schulhoff and Michael Ilie and Nishant Balepur and Konstantine Kahadze and Amanda Liu and Chenglei Si and Yinheng Li and Aayush Gupta and HyoJung Han and Sevien Schulhoff and Pranav Sandeep Dulepet and Saurav Vidyadhara and Dayeon Ki and Sweta Agrawal and Chau Pham and Gerson Kroiz and Feileen Li and Hudson Tao and Ashay Srivastava and Hevander Da Costa and Saloni Gupta and Megan L. Rogers and Inna Goncearenco and Giuseppe Sarli and Igor Galynker and Denis Peskoff and Marine Carpuat and Jules White and Shyamal Anadkat and Alexander Hoyle and Philip Resnik},
      year={2025},
      eprint={2406.06608},
      archivePrefix={arXiv},
      primaryClass={cs.CL},
      url={https://arxiv.org/abs/2406.06608}, 
}

@misc{wang2023rolellm,
      title={RoleLLM: Benchmarking, Eliciting, and Enhancing Role-Playing Abilities of Large Language Models}, 
      author={Zekun Moore Wang and Zhongyuan Peng and Haoran Que and Jiaheng Liu and Wangchunshu Zhou and Yuhan Wu and Hongcheng Guo and Ruitong Gan and Zehao Ni and Jian Yang and Man Zhang and Zhaoxiang Zhang and Wanli Ouyang and Ke Xu and Stephen W. Huang and Jie Fu and Junran Peng},
      year={2024},
      eprint={2310.00746},
      archivePrefix={arXiv},
      primaryClass={cs.CL},
      url={https://arxiv.org/abs/2310.00746}, 
}

@misc{liu2025temperature,
      title={Exploring the Impact of Temperature on Large Language Models:Hot or Cold?}, 
      author={Lujun Li and Lama Sleem and Niccolo' Gentile and Geoffrey Nichil and Radu State},
      year={2025},
      eprint={2506.07295},
      archivePrefix={arXiv},
      primaryClass={cs.CL},
      url={https://arxiv.org/abs/2506.07295}, 
}

@misc{looping2025reasoning,
      title={Wait, Wait, Wait... Why Do Reasoning Models Loop?}, 
      author={Charilaos Pipis and Shivam Garg and Vasilis Kontonis and Vaishnavi Shrivastava and Akshay Krishnamurthy and Dimitris Papailiopoulos},
      year={2025},
      eprint={2512.12895},
      archivePrefix={arXiv},
      primaryClass={cs.LG},
      url={https://arxiv.org/abs/2512.12895}, 
}

\end{document}